%% file: main.tex
\documentclass[11pt,a4paper]{article}
\usepackage[hyperref]{acl2019}
\usepackage{times}
\usepackage{latexsym}
\usepackage{comment}
\usepackage{graphicx}
\usepackage{comment}
\usepackage{enumitem}
\usepackage[utf8]{inputenc}

\usepackage{multirow}
\usepackage{sidecap}
\usepackage{amsthm}
\usepackage{amsmath}
\usepackage{amssymb}
\usepackage{amsfonts}
\usepackage{verbatim} 
\usepackage{url}
\usepackage{pifont}
\usepackage{eqparbox}
\usepackage{arydshln} 
\usepackage{bigstrut}
\usepackage[linewidth=1pt]{mdframed}
\usepackage{framed}
\usepackage{float}
\usepackage{booktabs}
\usepackage{url}
\usepackage[colorinlistoftodos,prependcaption,textsize=tiny]{todonotes}

\usepackage[normalem]{ulem}

\aclfinalcopy 

\definecolor{DarkRed}{RGB}{130,25,0}

\newcommand{\qiang}[1]{\textcolor{violet}{#1}}
\newcommand{\qiangcomment}[1]{\textcolor{violet}{\footnotesize{[\textit{QN: #1}]}}}
\newcommand{\ben}[1]{\textcolor{orange}{\footnotesize{[\textit{BZ: #1}]}}}
\newcommand{\benchange}[1]{\textcolor{orange}{[#1]}}

\usepackage{todonotes}
\newcommand{\bert}{\emph{BERT}}
\newcommand{\ignore}[1]{}

\newcommand{\namecite}[1]{\citeauthor{#1}~\shortcite{#1}}

\newcommand{\datasetname}{\textsc{McTaco}}
\usepackage{balance}
\title{
\vspace*{-0.5in}
{{\small \hfill EMNLP'19}\\
\vspace*{.25in}} 
\emph{``Going on a vacation''} takes longer than \emph{``Going for a walk''}: \\ 
A Study of Temporal Commonsense Understanding 
}

\author{Ben Zhou,$^1$ Daniel Khashabi,$^2$
Qiang Ning,$^{3}$ Dan Roth$^1$ \\
{ \normalsize  $^1$University of Pennsylvania, \;  $^2$Allen Institute for AI, \; $^3$University of Illinois at Urbana-Champaign } \\
{\tt \footnotesize \{xyzhou,danroth\}@cis.upenn.edu \; danielk@allenai.org \; qning2@illinois.edu}
}

\usepackage{lipsum}

\newcommand\blfootnote[1]{%
  \begingroup
  \renewcommand\thefootnote{}\footnote{#1}%
  \addtocounter{footnote}{-1}%
  \endgroup
}

\newcommand{\setOf}[1]{\left\lbrace #1 \right\rbrace}

\newcommand{\taco}[0]{\includegraphics[width=.028\textwidth]{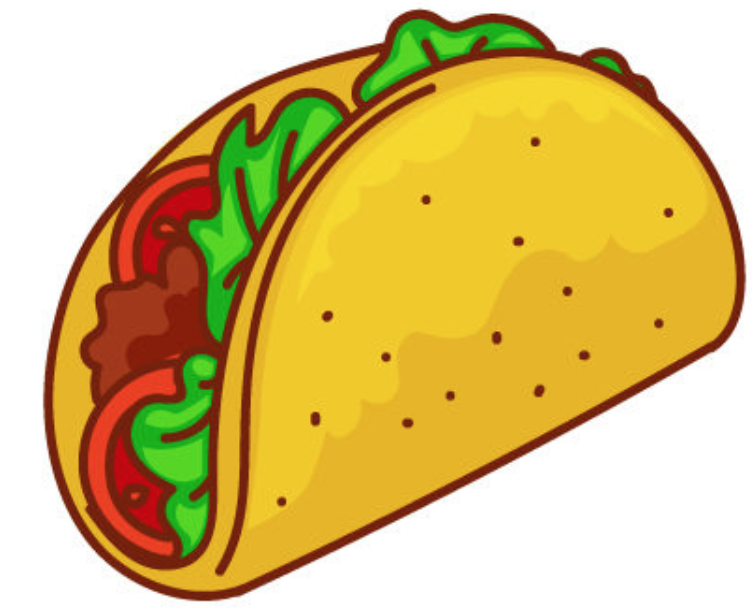}}

\date{}
\begin{document}
\maketitle
\input{abstract.tex}
\input{intro.tex}

\input{related.tex}

\input{tacoqa.tex}

\input{experiment.tex}
\input{conclusion.tex}

\section*{Acknowledgements}
This research is supported by a grant from the Allen Institute for Artificial Intelligence (allenai.org) and by contract HR0011-18-2-0052 and HR0011-15-C-0113 with the US Defense Advanced Research Projects Agency (DARPA).
Approved for Public Release, Distribution Unlimited. The views expressed are those of the authors and do not reflect the official policy or position of the Department of Defense or the U.S. Government.


\bibliography{ccg}
\bibliographystyle{acl_natbib}
\input{appendix.tex}
\end{document}

%% file: abstract.tex
\begin{abstract}

Understanding time is crucial for understanding events expressed in natural language. Because people rarely say the obvious, it is often necessary to have commonsense knowledge about various temporal aspects of events, such as duration, frequency, and temporal order. However, this important problem has so far received limited attention. 
This paper systematically studies this {\em temporal commonsense} problem.
Specifically, we define five classes of temporal commonsense, and 
use crowdsourcing to develop a new dataset, \datasetname~\taco,  that serves as a test set for this task.
We find that the best current methods used on \datasetname{} are still far behind human performance, by about 20\%, and discuss several directions for improvement.
We hope that the new dataset and our study here can foster more future research on this topic.\footnote{
The dataset, annotation interfaces, guidelines, and qualification tests are available at: { \url{https://cogcomp.seas.upenn.edu/page/publication_view/882}}.
}

\end{abstract}

%% file: intro.tex
\section{Introduction}
\begin{figure}[h!]
    \centering
    \includegraphics[width=0.48\textwidth,trim=0.3cm 0cm 0cm 0cm]{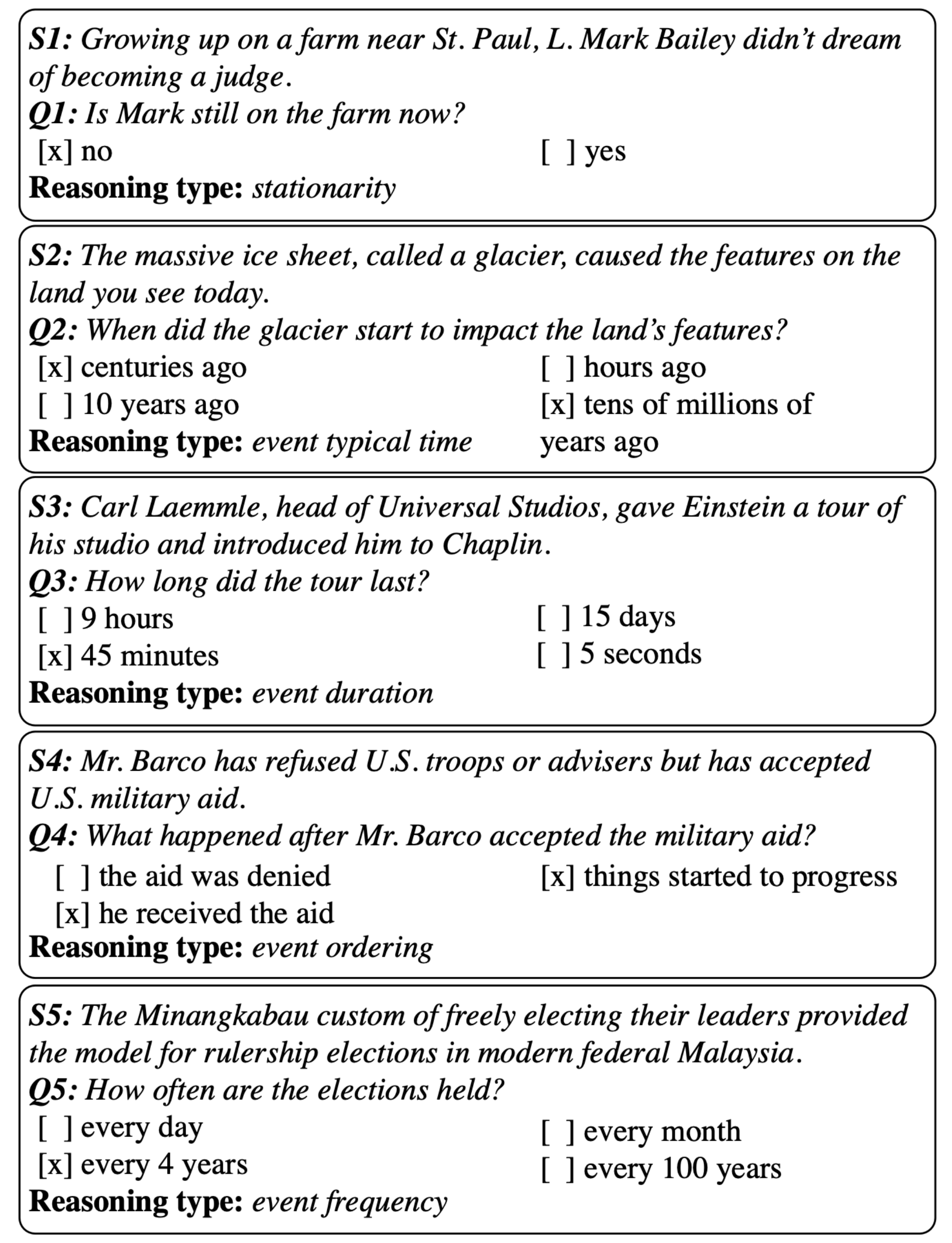}
    \caption{\small Five types of temporal commonsense in \datasetname. Note that a question may have \emph{multiple} correct answers. \ignore{Correct answers marked by `x'.}}
    \label{fig:intro:example}
\end{figure}

Natural language understanding requires the ability to reason with {\em commonsense} knowledge~\cite{schubert2002can,davis2014representations}, 
and the last few years have seen significant amount of work in this direction
(e.g., \citet{ZRDV17,BauerWaBa18,TDGYBC18}).
This work studies a specific type of commonsense: {\em temporal commonsense}. For instance, given two events \emph{``going on a vacation''} and \emph{``going for a walk,''} most humans would know that a vacation is typically longer and occurs less often than a walk, but it is still challenging for computers to understand and reason about temporal commonsense.

\blfootnote{* This work was done while the second author was affiliated with the University of Pennsylvania.}

Temporal commonsense has received limited attention so far. \textbf{Our first contribution} is that, to the best of our knowledge, we are the first to systematically study and quantify performance on a range of temporal commonsense phenomena. Specifically, we consider five temporal properties: \emph{duration} (how long an event takes),  \emph{temporal ordering} (typical order of events), \emph{typical time} (when an event happens), \emph{frequency} (how often an event occurs), and \emph{stationarity} (whether a state holds for a very long time or indefinitely). Previous work has investigated some of these aspects, either explicitly or implicitly (e.g., duration \cite{divyekhilnani2011using,Williams12} and ordering \cite{ChklovskiPa04,NWPR18}), but none of them have defined or studied all aspects of temporal commonsense in a unified framework. \citet{kozareva2011learning} defined a few temporal aspects to be investigated, but failed to quantify performances on these phenomena. 

Given the lack of evaluation standards and datasets for temporal commonsense, \textbf{our second contribution} is the development 
of a new dataset dedicated for it, \datasetname{} (short for \textbf{m}ultiple \textbf{c}hoice \textbf{t}empor\textbf{a}l \textbf{co}mmon-sense).
\datasetname{} is constructed via crowdsourcing with guidelines designed meticulously
to guarantee its quality.
When evaluated on \datasetname{}, a system receives a \emph{sentence} providing context information, a \emph{question} designed to require temporal commonsense knowledge, and multiple \emph{candidate answers} (see Fig.~\ref{fig:intro:example}; note that in our setup, more than one candidate answer can be plausible).
We design the task as a binary classification: determining whether a candidate answer
is \emph{plausible} according to human commonsense, since there is no \emph{absolute} truth
here.  This is aligned with other efforts that have posed commonsense as the choice of plausible alternatives~\cite{roemmele2011choice}. 
The high quality of the resulting dataset (shown in \S\ref{sec:exp}) also makes us believe that the notion of plausibility here is robust.

\ignore{\benchange{\sout{The task is a binary classification of whether the candidate answer is a plausible a response according to human's commonsense. }}}



\textbf{Our third contribution} is that, using \datasetname{} as a testbed, we study the temporal commonsense understanding of the best existing NLP techniques, including \emph{ESIM} \cite{Chen-Qian:2017:ACL}, 
\bert{} \cite{devlin2018bert} 
and their variants.
Results in \S\ref{sec:exp} show that, despite a significant improvement over random-guess baselines, the best existing techniques are still far behind human performance on temporal commonsense understanding,
indicating the need for further research in order to improve the currently limited capability to capture temporal semantics.

%% file: related.tex
\section{Related Work}
\label{sec:related_work}

Commonsense has been a very popular topic in recent years and existing NLP works have mainly investigated the acquisition and evaluation of commonsense in the physical world, including but not limited to, size, weight, and strength \cite{forbes2017verb}, roundness and deliciousness \cite{yang2018extracting}, and intensity \cite{CWPAC18}. In terms of ``events" commonsense, \citet{RSASC18} investigated the intent and reaction of participants of an event, and \citet{zellers2018swag} tried to select the most likely subsequent event.
To the best of our knowledge, no earlier work has focused on \emph{temporal} commonsense, although it is critical for event understanding.
For instance, \citet{NingWuRo18} argues that resolving ambiguous and implicit mentions of event durations in text (a specific kind of temporal commonsense)
is necessary to construct the timeline of a story.

There have also been many works trying to understand time in natural language but not necessarily the commonsense understanding of time. Most recent works include the extraction and normalization of temporal expressions~\cite{strotgen2010heideltime,LADZ14}, temporal relation extraction \cite{NingFeRo17,NZFPR18}, and timeline construction \cite{LeeuwenbergMo18}. Among these, some works are implicitly on temporal commonsense, such as event durations \cite{Williams12,VempalaBlPa18}, typical temporal ordering \cite{ChklovskiPa04,NFWR18,NWPR18}, and script learning (i.e., what happens next after certain events) \cite{GranrothCl16,LiDiLi18}. 
However, existing works have not studied all five types of temporal commonsense in a unified framework as we do here, nor have they developed datasets for it.

Instead of working on each individual aspect of temporal commonsense, we formulate the problem as a machine reading comprehension task in the format of selecting plausible responses with respect to natural language queries. 
This relates our work to a large body of work on question-answering, an area that has seen significant progress in the past few years
\cite{Clark2018ThinkYH,ostermann2018semeval,merkhofer2018mitre}. This area, however, has mainly focused on {\em general} natural language comprehension tasks, while we tailor it to test a {\em specific} reasoning capability, which is temporal commonsense. 

\ignore{
\paragraph{Machine Reading Comprehension.}
This area has witnessed many recent datasets in the past few years. 
Many of the proposals provide general natural language comprehension tasks, without tailoring it to any specific reasoning ability.   
To name a few, 
\namecite{Clark2018ThinkYH,ostermann2018semeval,merkhofer2018mitre} are among such works. 
While these are 
helpful to measure general language 
understanding,
they are not very useful when evaluating narrow 
aspects of language. 
Our reading comprehension dataset, while being naturally generated by 
crowdsourcers,
it is 
focused on class of language understanding problems (various properties of time)
and contains 
granular annotations for the required skills. 
}

%% file: tacoqa.tex
\begin{table}[]
\centering
\footnotesize
\resizebox{0.45\textwidth}{!}{
\begin{tabular}{lcc}
\toprule
\multicolumn{2}{l}{Measure}                            & Value         \\
\cmidrule(r){1-2}  \cmidrule(r){3-3}
\multicolumn{2}{l}{\# of unique questions}             & 1893              \\
\multicolumn{2}{l}{\# of unique question-answer pairs}             & 13,225              \\
\multicolumn{2}{l}{avg. sentence length}            & 17.8              \\
\multicolumn{2}{l}{avg. question length}               & 8.2              \\
\multicolumn{2}{l}{avg. answer length}                 & 3.3              \\
\midrule
Category                           & \# questions & avg \# of candidate \\
\cmidrule(r){1-1} \cmidrule(r){2-2} \cmidrule(r){3-3}
\emph{event frequency}                    & 433 & 8.5 \\
\emph{event duration}                     & 440 & 9.4 \\
\emph{event stationarity}                 & 279 & 3.1 \\
\emph{event ordering}                     & 370 & 5.4 \\
\emph{event typical time}                 & 371 & 6.8 \\
\bottomrule
\end{tabular}
}
\caption{\small Statistics of \datasetname. }
    \label{tab:statistics}
\end{table}

\section{Construction of \datasetname}
\label{sec:tacoqa}

\ignore{
In the next steps we describe  a
multi-step crowdsourcing scheme, 
resulting from 
 detailed analysis and substantial refinements after multiple pilots studies. 
}


\datasetname{} is comprised of $13k$ tuples, in the form of {\em (sentence, question, candidate answer)}; please see examples in Fig.~\ref{fig:intro:example} for the five phenomena studied here and Table~\ref{tab:statistics} for basic statistics of it. 
The sentences in those tuples are 
randomly selected from MultiRC~\cite{MultiRC2018} 
(from each of its 9 domains). 
For each sentence, we use crowdsourcing on Amazon Mechanical Turk to collect questions and candidate answers (both correct and wrong ones).
To ensure the quality of the results, we limit the annotations to native speakers and use qualification tryouts.

\paragraph{Step 1: Question generation.}
We first ask crowdsourcers to generate questions, given a sentence.
To produce questions that need temporal commonsense to answer, we require that a valid question: (a) should ask about one of the five temporal phenomena we defined earlier, and (b) should not be solved simply by a word or phrase from the original sentence.
We also require crowdsourcers to provide a correct answer for each of their questions, which on one hand gives us a positive candidate answer, and on the other hand ensures that the questions are answerable at least by themselves.

\paragraph{Step 2: Question verification.}
We further ask another two crowdsourcers to check the questions generated in Step~1, i.e., (a) whether the two requirements are satisfied and (b) whether the question is grammatically and logically correct.
We retain only the questions where the two annotators unanimously agree with each other and the decision generated in Step~1.
For valid questions, we continue to ask crowdsourcers to give one correct answer and one incorrect answer, which we treat as a seed set to 
automatically generate new candidate answers in the next step.

\paragraph{Step 3: Candidate answer expansion.}
Until this stage, we have collected a small set of candidate answers (3 positive and 2 negative) for each question.\footnote{One positive answer from Step~1; one positive and one negative answer from each of the two annotators in Step~2.} We automatically expand this set in three ways. First, we use a set of rules to extract numbers and quantities (``2", ``once") and temporal terms (e.g. ``a.m.", ``1990", ``afternoon", ``day"), and then randomly perturb them based on a list of temporal units (``second''), adjectives (``early''), points ( ``a.m.'') and adverbs (``always''). 
Examples are ``2 a.m." $\rightarrow$ ``3 p.m.", ``1 day" $\rightarrow$ ``10 days", ``once a week"$\rightarrow$ ``twice a month" (more details in the appendix). 

Second, we mask each individual token in a candidate answer (one at a time) and use \bert{}~\cite{devlin2018bert} to predict replacements for each missing term; we rank those predictions by the confidence level of \bert{} and keep the top three.
\ignore{ones \qiangcomment{how many?}.}

Third, for those candidates that represent events, 
the previously-mentioned token-level perturbations rarely lead to interesting and diverse set of candidate answers. Furthermore, it
may lead to invalid phrases (e.g., ``he left the house'' $\rightarrow$ ``he walked the house''.) Therefore, to perturb such candidates, we create a pool of $60k$ event phrases using PropBank~\cite{kingsbury2002treebank}, and perturb 
the candidate answers to be the most similar ones extracted by an information retrieval (\emph{IR}) system.\footnote{\url{www.elastic.co}} 
This not only guarantees that all candidates are properly phrased, it also leads to more diverse perturbations. 

We apply the  above three techniques on non-``event'' candidates sequentially, in the order they were explained, to expand the candidate answer set to 20 candidates per question. 
A perturbation technique is used, as long as the pool of candidates is still less than 20.
Note there are both correct and incorrect answers in those candidates.



\paragraph{Step 4: Answer labeling.}
In this step, each ({\em sentence, question, answer}) tuple produced earlier is labeled by 4 crowdsourcers, with three options: ``likely'', ``unlikely'', or ``invalid'' (sanity check for valid tuples).\footnote{We use the name ``(un)likely'' because commonsense decisions can be naturally ambiguous and subjective.} Different annotators may have different interpretations, yet we ensure label validity through high agreement.
A tuple is kept only if
all 4 annotators agree on ``likely'' or ``unlikely''.
The final statistics of \datasetname{} is in Table~\ref{tab:statistics}.

%% file: experiment.tex
\section{Experiments}
\label{sec:exp}
We assess the quality of our dataset through human annotation, and evaluate a couple of baseline systems. 
We create a uniform split of 30\%/70\% of the data to dev/test.
The rationale behind this split is that, a successful system has to bring in a huge amount of world knowledge and derive commonsense understandings {\em prior} to the current task evaluation. We therefore believe that it is not reasonable
to expect a system to be {\em trained} solely on this data, and we think of the development data as only providing a {\em definition} of the task. Indeed, the gains from our development data are marginal after a certain number of  training instances. 
This intuition is studied and verified in Appendix~\ref{appendix:performance:vs:size}.

\paragraph{Evaluation metrics.}
Two question-level metrics are adopted in this work: exact match (\emph{EM}) and {\em F1}. 
For a given candidate answer $a$ that belongs to a question $q$, let $f(a; q) \in \{0, 1\}$ denote the correctness of the prediction made by a fixed system (1 for correct; 0 otherwise). Additionally, let $D$ denote the collection of questions in our evaluation set.
$$EM \triangleq \frac{\sum_{q\in D} \prod_{a \in q}  f(a; q) }{ | \setOf{ q\in D }| }$$
The recall for each question $q$ is:
$$
R(q) = \frac{ \sum_{a \in q} \left[f(a; q) = 1 \right] \wedge \left[ a \text{ is ``likely" } \right] }{ |\setOf{a \text{ is ``likely" } \wedge a \in q }| }
$$
Similarly, $P(q)$ and $F1(q)$ are defined. The aggregate $F1$ (across the dataset $D$) is the macro average of question-level $F1$'s:
$$F1 \triangleq  \frac{\sum_{q \in D}F1(q)}{ | \setOf{ q\in D }| }$$

\emph{EM} measures how many questions a system is able to correctly label all candidate answers, while {\em F1} is more relaxed and measures the average overlap between one's predictions and the ground truth. 

\paragraph{Human performance.}
An expert annotator also worked on \datasetname{} to gain a better understanding of the human performance on it.
The expert answered 100 questions (about 700 ({\em sentence, question, answer}) tuples) randomly sampled from the test set, and could only see a single answer at a time, with its corresponding question and sentence. 

\paragraph{Systems.} 
We use two state-of-the-art systems in machine reading comprehension for this task: \emph{ESIM}~\cite{Chen-Qian:2017:ACL} and \bert{}~\cite{devlin2018bert}.
\emph{ESIM} is an effective neural model on natural language inference. We initialize the word embeddings in {\em ESIM} via either \emph{GloVe}~\cite{pennington2014glove} or \emph{ELMo}~\cite{peters2018deep} to demonstrate the effect of pre-training.
\bert{} is a state-of-the-art contextualized representation used for a broad range of tasks 
\ignore{\qiangcomment{I don't like ``high-level'': either delete it or be more specific. Simply saying ``high-level'' is not helpful for our point here.}}. 
We also add unit normalization to \bert{}, which extracts and converts temporal expressions in candidate answers to their most proper units. For example, ``30 months'' will be converted to ``2.5 years".
To the best of our knowledge, there are no other available systems for the ``stationarity'', ``typical time'', and ``frequency'' phenomena studied here. As for ``duration'' and ``temporal order'', there are existing systems (e.g., \citet{VempalaBlPa18,NWPR18}), but they cannot be directly applied to the setting in \datasetname{} where the inputs are natural languages.

\paragraph{Experimental setting.}
In both \emph{ESIM} baselines, we model the process as a sentence-pair classification task, following the \emph{SNLI} setting in AllenNLP.\footnote{\url{https://github.com/allenai/allennlp}}
In both versions of \emph{BERT}, we use the same sequence pair classification model and the same parameters as in \emph{BERT}'s \emph{GLUE} experiments.\footnote{\url{https://github.com/huggingface/pytorch-pretrained-BERT}}
A system receives two elements
at a time: (a) the concatenation of the sentence and question, and (b) the answer. 
The system makes a 
binary prediction on each instance, ``likely" or ``unlikely".



\begin{table}[]
    \centering
    \small 
    \begin{tabular}{ccc}
        \toprule 
         System & \emph{F1} & \emph{EM}  \\ 
         \cmidrule(lr){1-1} \cmidrule(lr){2-2} \cmidrule(lr){3-3} 
         Random  & 36.2 & 8.1 \\  
         Always Positive & 49.8 & 12.1  \\
         Always Negative & 17.4 & 17.4 \\
         \midrule
         \emph{ESIM + GloVe} & 50.3 & 20.9 \\
         \emph{ESIM + ELMo} & 54.9 & 26.4\\
         \emph{BERT} & 66.1 & 39.6  \\
         \emph{BERT + unit normalization} & \textbf{69.9} & \textbf{42.7}  \\ 
         \midrule 
         Human & 87.1 & 75.8  \\ 
        \bottomrule 
    \end{tabular}
    \caption{\small Summary of the performances for different baselines. All numbers are in percentages. 
    }
    \label{tab:performance}
\end{table}

\paragraph{Results and discussion.}
Table~\ref{tab:performance} compares native baselines, {\em ESIM}, \bert{} and their variants on the entire test set of \datasetname{}; it also shows human performance on the subset of 100 questions.\footnote{\emph{BERT + unit normalization} scored $F1=72,EM=45$ on this subset, which is only slightly different from the corresponding number on the entire test set.}
The system performances reported are based on default random seeds, and we observe a maximum standard error~\footnote{\url{https://en.wikipedia.org/wiki/Standard_error}} of 0.8 from 3 runs on different seeds across all entries.
We can confirm the good quality of this dataset based on the high performance of human annotators. \emph{ELMo} and \bert{} improve naive baselines by a large margin, indicating that 
a notable amount of
commonsense knowledge has been acquired via pre-training. However, even \bert{} still falls far behind human performance, indicating the need of further research. \footnote{RoBERTa~\cite{liu2019roberta},
a more recent language model
that was released after this paper's submission, achieves $F1=72.3, EM=43.6.$}



\begin{figure}
    \centering
    \includegraphics[scale=0.16,trim=1.6cm 0cm 0cm 1.5cm, clip=false]{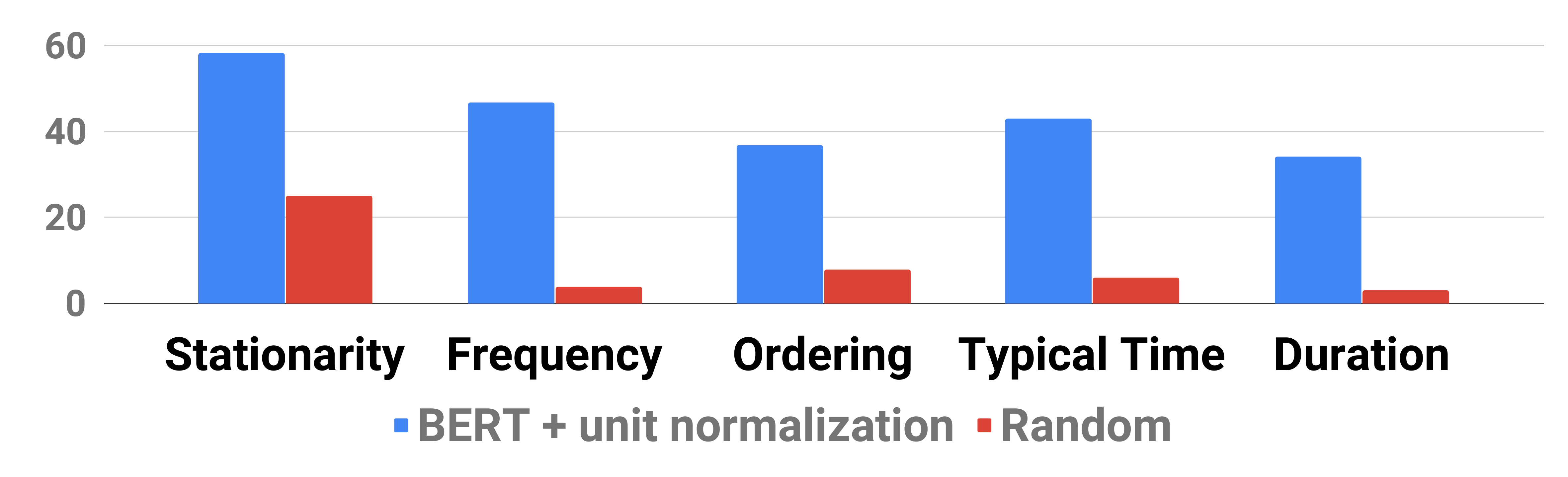}
    \caption{\small \emph{EM} scores of \emph{BERT + unit normalization} per temporal reasoning category comparing to the random-guess baseline.}
    \label{fig:bert-per-category}
\end{figure}
\ignore{
In this section we provide further 
intuition into our results. 
A major takeaway from the experiments, as expected, is the importance of \emph{contextual pre-training} for a task like ours, which involves much commonsense understanding (unlike tasks where answers are part of the input) \qiangcomment{not sure about the ``unlike'' part}. The best system in our results strongly relies on BERT. Additionally, the combination of \emph{ESIM} with \emph{ELMo} (a contextual pre-trained representation), is much stronger than using \emph{GloVe} as the underlying representations. \todo{Two 'buts' very close to each other; also not sure what "sensitive to units" means. } This indicates the importance of using large amount of world knowledge in this task. \qiangcomment{a bit wordy; can just say ``the significant improvement brought by contextualized pre-training such as \emph{BERT} and \emph{ELMo} indicates that a significant portion of commonsense knowledge is actually acquired via pre-training.''}
}

\newcommand{\cmark}{\ding{51}}%
\newcommand{\xmark}{\ding{55}}%

We know that \bert{}, as a language model, is good at associating surface forms (e.g. associating ``sunrise'' with ``morning'' since they often co-occur), but may be 
brittle with respect to variability of temporal mentions. 

Consider the following example (the correct answers are indicated with \cmark and \bert\ selections are \underline{underlined}.)  
This is an example of \bert{} correctly associating a given event with ``minute" or``hour"; however, it fails to distinguish between ``1 hour'' (a ``likely'' candidate)  and ``9 hours'' (an ``unlikely'' candidate).
\begin{framed}
\small \noindent
\textbf{P:} \emph{Ratners's chairman, Gerald Ratner, said the deal remains of "substantial benefit to Ratners."}\\
\textbf{Q:} \emph{How long did the chairman speak?}  \\ 
\; \cmark(a) \underline{30 minutes} \hspace{0.8cm}  \cmark (b) \underline{1 hour} \\ 
\; \xmark (c) \underline{9 hours}  \hspace{1.35cm}  \xmark (d) twenty seconds  
\end{framed}

This shows that \bert{} does not infer a range of true answers; it instead associates discrete terms and decides individual options, which may not be the best way to handle temporal units that involve numerical values.

{\em BERT+unit normalization} is used to address this issue, but results show that it is still poor compared to human. This indicates that the information acquired by \bert{} is still far from solving temporal commonsense.


Since exact match (EM) is a stricter metric, it is consistently lower than $F1$ in Table~\ref{tab:performance}. 
For an ideal system, the gap between EM and $F1$ should be small (humans only drop 11.3\%.)
However, all other systems drop by almost 30\% from $F1$ to EM, possibly 
another piece of evidence that they only associate surface forms instead of using one representation for temporal commonsense to classify all candidates.
A curious reader might ask why the human performance on this task as shown in Table~\ref{tab:performance} is not 100\%.
This is expected because commonsense is what {\em most} people agree on, so any {\em single} human could disagree with the gold labels in \datasetname{}. Therefore, we think the human performance in Table~\ref{tab:performance} from a single evaluator actually indicates the good quality of \datasetname{}.

The performance of \emph{BERT+unit normalization} is not uniform across different categories (Fig.~\ref{fig:bert-per-category}), which could be due to the different nature or quality of data for those temporal phenomena. For example, as shown in Table~\ref{tab:statistics}, ``stationarity'' questions have much fewer candidates and a higher random baseline. 

%% file: conclusion.tex
\section{Conclusion}
\label{sec:conclusion}
This work has focused on temporal commonsense.
We define five categories of questions that require temporal commonsense and develop a novel crowdsourcing scheme to generate \datasetname{}, a high-quality dataset for this task. 
We use \datasetname{} to probe the capability of systems on temporal commonsense understanding. 
We find that
systems equipped with state-of-the-art language models such as {\em ELMo} and \bert{} are still far behind humans, thus motivating future research in this area. Our analysis sheds light on the capabilities as well as limitations of current models. 
We hope that this study will inspire further research on temporal commonsense. 


%% file: appendix.tex
\clearpage

\appendix
\onecolumn
\section{Supplemental Material}
\label{sec:supplemental}

\subsection{Perturbing Candidate Answers}
\label{appendix:question:perturbations}
Here we provide a few missing details from \emph{Step 3} of our annotations (Section 3). In particular, we create collections of common temporal expressions (see Table~\ref{tab:temporal:expressions}) to detect whether the given candidate answer contains a temporal expression or not. If a match is found within this list, we use the mappings to create perturbations of the temporal expression. 

\begin{table}[h]
    \centering
\noindent
\resizebox{\textwidth}{!}{
{
\footnotesize
\begin{tabular}{|c|c|c|c|c|}
    \hline 
     Adjectives & Frequency & Period & Typical time & Units  \\
     \hline 
     early:late	& always:sometimes:never	& night:day	& now:later	& second:hour:week:year \\ 
late:early	& occasionally:always:never	& day:night	& today:yesterday	& seconds:hours:weeks:years \\ 
morning:late night	& often:rarely	& 	& tomorrow:yesterday	& minute:day:month:century \\ 
night:early morning	& usually:rarely	& 	& tonight:last night	& minutes:days:months:centuries \\ 
evening:morning	& rarely:always	& 	& yesterday:tomorrow	& hour:second:week:year \\ 
everlasting:periodic	& constantly:sometimes	& 	& am:pm	& hours:seconds:weeks:years \\ 
initial:last	& never:sometimes:always	& 	& pm:am	& day:minute:month:century \\ 
first:last	& regularly:occasionally:never	& 	& a.m.:p.m.	& days:minutes:months:centuries \\ 
last:first	& 	& 	& p.m.:a.m.	& week:second:hour:year \\ 
overdue:on time	& 	& 	& afternoon:morning	& weeks:seconds:hours:years \\ 
belated:punctual	& 	& 	& morning:evening	& month:minute:day:century \\ 
long-term:short-term	& 	& 	& night:morning	& months:minutes:days:centuries \\ 
delayed:early	& 	& 	& after:before	& year:second:hour:week \\ 
punctual:belated	& 	& 	& before:after	& years:seconds:hours:weeks \\ 
	& 	& 	& 	& century:minute:day:month \\ 
	& 	& 	& 	& centuries:minutes:days:months \\ 
     \hline 
\end{tabular}
}
}
    \caption{Collections of temporal expressions used in creating perturbation of the candidate answers. Each mention is grouped with its variations (e.g., ``first'' and ``last'' are in the same set). }
    \label{tab:temporal:expressions}
\end{table}




\subsection{Performance as a function of training size}
\label{appendix:performance:vs:size}
An intuition that we stated is that, the task at hand requires a successful model to bring in external world knowledge beyond what is observed in the dataset; since for a task like this, it is unlikely to compile an dataset which covers all the possible events and their attributes.
In other words, the ``traditional'' supervised learning alone (with no pre-training or external training) is unlikely to succeed. 
A corollary to this observation is that, tuning a pre-training system (such as BERT~\cite{devlin2018bert}) likely requires very little supervision.   

\begin{figure}[h]
    \centering
    \includegraphics[scale=0.4,trim=0cm 0cm 0cm 0cm, clip=false]{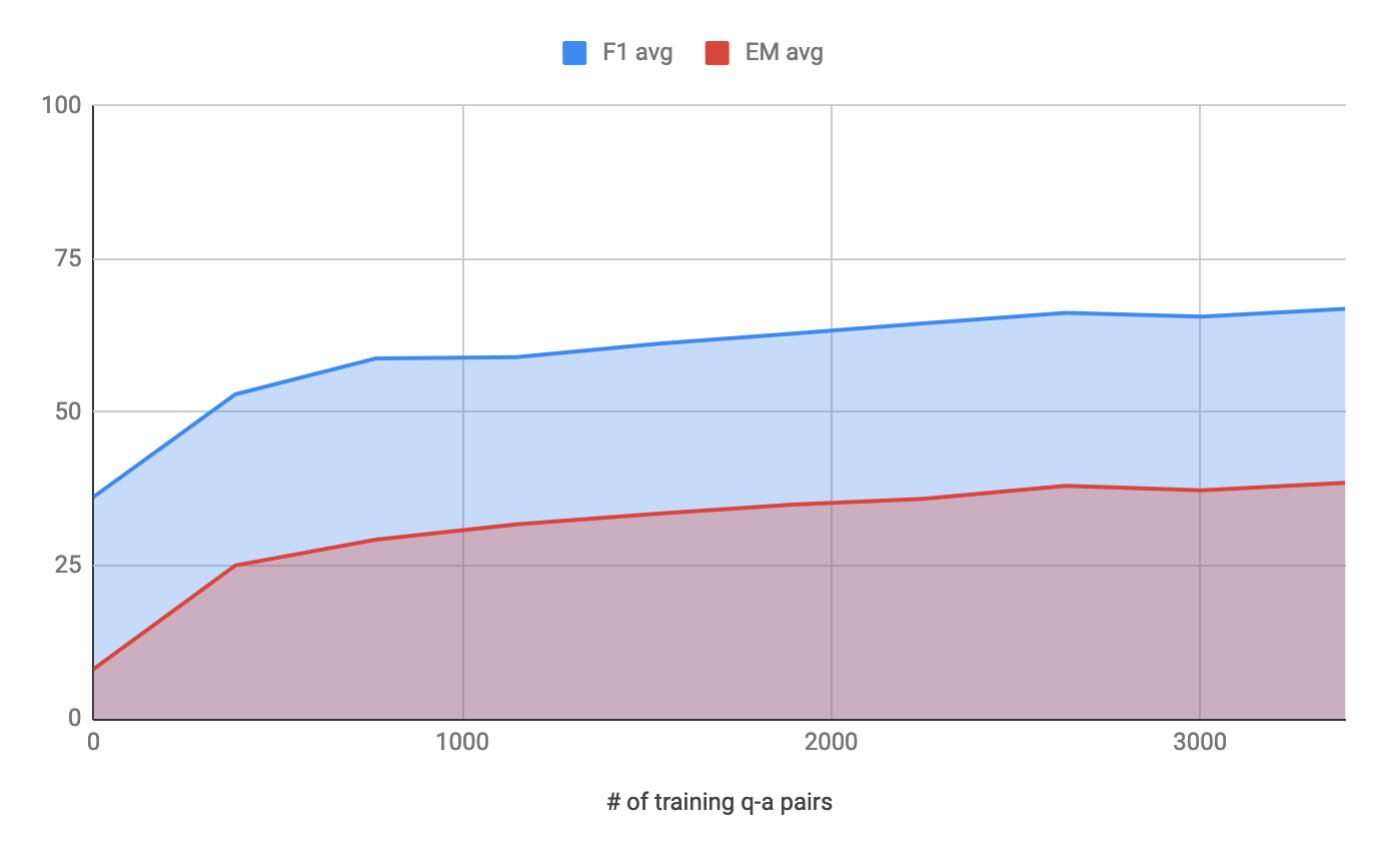}
    \caption{Performance of supervised algorithm (BERT; Section~{4}) as function of various sizes  of observed training data. When no training data provided to the systems (left-most side of the figure), the performance measures amount to random guessing.}
    \label{fig:performance:vs:training:size}
\end{figure}

We plot the performance change, as a function of number of instances observed in the training time (Figure~\ref{fig:performance:vs:training:size}). 
Each point in the figure share the same parameters and averages of 5 distinct trials over different random sub-samples of the dataset.
As it can be observed, the performance plateaus after about $2.5k$ question-answer pairs (about 20\% of the whole datasets).
This verifies the intuition that systems can rely on a relatively small amount of supervision to tune to task, if it models the world knowledge through pre-training. Moreover, it shows that trying to make improvement through getting more labeled data is costly and impractical.

\subsection{Annotation Interfaces}
\label{sec:supp:screenshots}

\ignore{
\todo{In Appendix A.3, there UI shows \$(sentence), \$(question), \$(answer), etc. This needs to be corrected.}
}
\begin{figure}[h!]
    \centering
    \includegraphics[width=0.7\columnwidth]{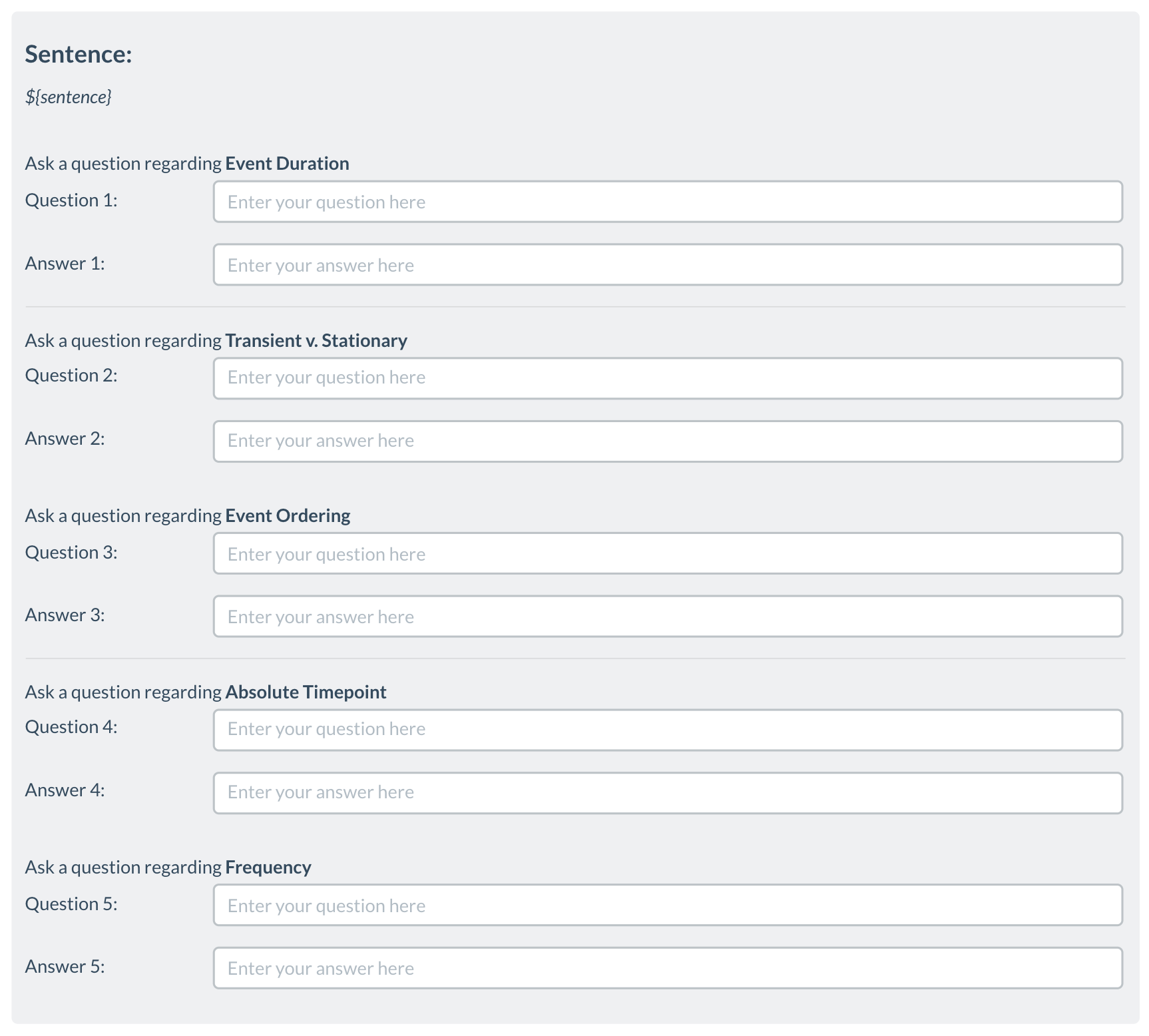}
    \caption{Step 1}
    \label{fig:screenshots}
\end{figure}

\begin{figure}[h!]
    \centering
    \includegraphics[width=0.7\columnwidth]{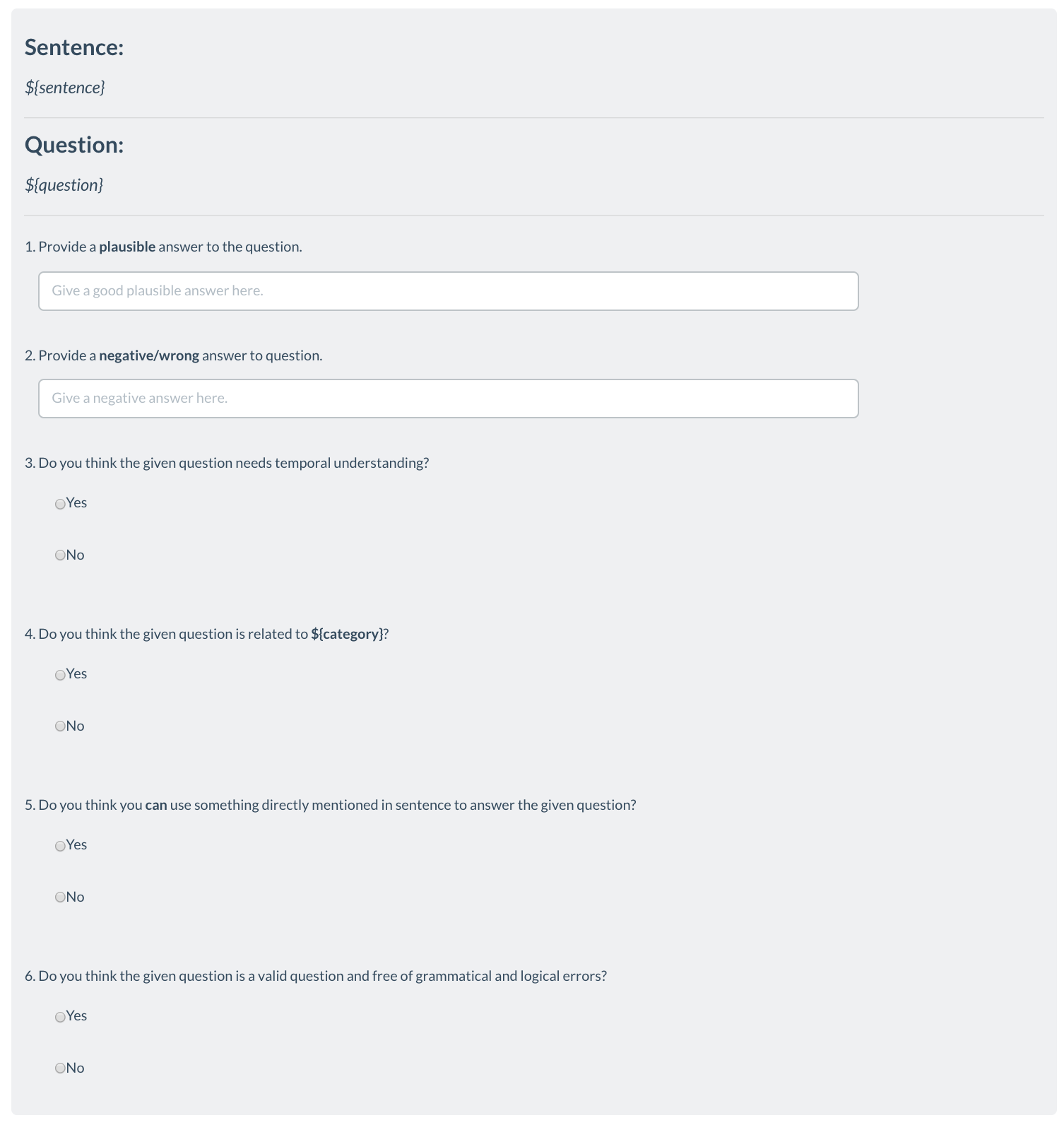}
    \caption{Step 2}
    \label{fig:screenshots}
\end{figure}

\begin{figure}[h!]
    \centering
    \includegraphics[width=0.7\columnwidth]{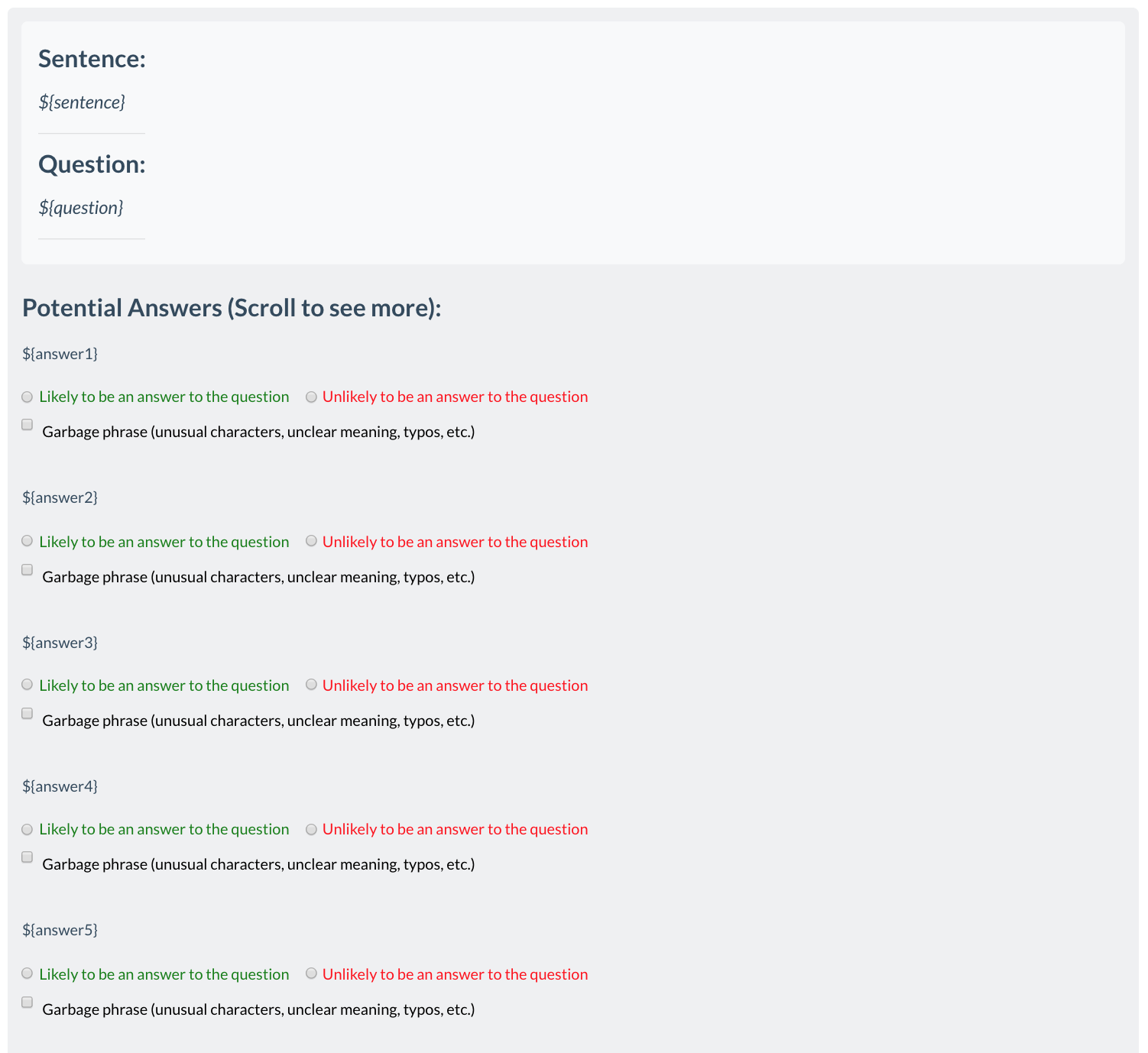}
    \caption{Step 3}
    \label{fig:screenshots}
\end{figure}